\documentclass[a4paper,twoside]{article}

\usepackage{epsfig}
\usepackage{subcaption}
\usepackage{calc}
\usepackage{amssymb}
\usepackage{amstext}
\usepackage{amsmath}
\usepackage{amsthm}
\usepackage{multicol}
\usepackage{pslatex}
\usepackage{apalike}
\usepackage{algorithm2e}
\usepackage[bottom]{footmisc}
\usepackage{xcolor}
\usepackage{url}
\usepackage{tikz}
\usepackage{multirow}

\usepackage{SCITEPRESS}     

\begin{document}

\title{The Effect of Belief Boxes and Open-mindedness on Persuasion}


\author{\authorname{Onur Bilgin\sup{1}, Abdullah As Sami\sup{1}, Sriram Sai Vujjini\sup{1} and John Licato\sup{1}}
\affiliation{\sup{1}Advancing Machine and Human Reasoning (AMHR) Lab, University of South Florida}}



\abstract{As multi-agent systems are increasingly utilized for reasoning and decision-making applications, there is a greater need for LLM-based agents to have something resembling propositional beliefs. One simple method for doing so is to include statements describing beliefs maintained in the prompt space (in what we'll call their ``belief boxes''). But when agents have such statements in belief boxes, how does it actually affect their behaviors and dispositions towards those beliefs? And does it significantly affect agents' ability to be persuasive in multi-agent scenarios? Likewise, if the agents are given instructions to be open-minded, how does that affect their behaviors? We explore these and related questions in a series of experiments.
Our findings confirm that instructing agents to be open-minded affects how amenable they are to belief change. We show that incorporating belief statements and their strengths influences an agent’s resistance to (and persuasiveness against) opposing viewpoints. Furthermore, it affects the likelihood of belief change, particularly when the agent is outnumbered in a debate by opposing viewpoints, i.e., \textit{peer pressure} scenarios. The results demonstrate the feasibility and validity of the belief box technique in reasoning and decision-making tasks.}

\onecolumn \maketitle \normalsize \setcounter{footnote}{0} \vfill

\section{\uppercase{Introduction}}
\label{sec:introduction}

Argumentation is an important component of the reasoning and decision-making process. It allows individuals to communicate viewpoints, justifications, and evidence. But there is a complex relationship between argumentation and belief. For example, individuals who hold beliefs strongly may be less receptive to arguments that go against those beliefs. 
Multi-agent large language model-based (LLM) systems have shown remarkable capabilities in various tasks such as problem-solving, decision-making, and reasoning \cite{qian2023communicative,li2023camel,xu2023language,xiong2023examining,xu2023towards}. But understanding the \textit{beliefs} of these agents (or the LLM-based analogue) is difficult, as the dispositions of agents are encoded in the distribution of weights in their neural architectures. It can therefore be convenient (in terms of control, configuration, and explainability) to include an agent's beliefs as explicit text that is provided in its input prompt. But how does the inclusion of beliefs in this way actually affect the individual and social behaviors of these agents? For example, in environments where multiple agents with differing beliefs interact, do belief boxes affect agents' abilities to persuade each other, or to influence each other via peer pressure?

We set out to explore answers to these questions. 
In this paper, we define a construct that modifies an agent's convictions about a natural-language proposition, leading the agent to adopt and defend it in a debate. We represent it with a \textit{belief box},\footnote{We borrow this term from philosophy of mind and cognitive science, where it often describes a common assumption of computationalism \cite{rescorla2019language,schiffer1981truth}.} a set of beliefs as propositions, each with a Likert scale strength value indicating the agent's confidence in that belief. 
Hereafter, references to agents' beliefs (e.g., aligned/misaligned or correct/incorrect) refer to the contents of the belief box, and we aim to evaluate whether the belief box aligns with the expected behavior of agents. Correct/incorrect refer to objective labels, while aligned/misaligned reflect beliefs created from human-annotated arguments in corresponding datasets. We explored beliefs in multi-agent systems, how the agents with beliefs interact with each other, how they evaluate and change their beliefs, and how persuasive the group dynamics are.

\begin{table*}[htb!]    
    \caption{Research questions, hypotheses, and results of our work in overview.}
    \centering
    \begin{tabular}{ l | l | l}
        \hline
        Research Question & Hypothesis & Results \\
        \hline
        RQ1: Does prompting LLMs to be & Prompting LLMs to be open-minded will & Supported \\  
        open-minded cause more belief change? & lead to more belief changes in their given & (BFI-2, \\
         & belief boxes. & Aporia) \S \ref{sec:validatingOpen-mindedness} \\
        \hline        
        RQ2: What are the effects of beliefs in a & Agents that believe a proposition $p$ will be & Supported \\
        belief box on the agent's persuasion ability? & more persuasive in arguing that $p$ is true. & (Aporia) \S \ref{sec:beliefboxPersuasiveness} \\
        \cline{2-3}
         & Agents that believe a proposition $\neg p$ will & Supported  \\
         & be less persuasive in arguing that $p$ is true. & (Aporia) \S \ref{sec:beliefboxPersuasiveness} \\
        \hline        
        RQ3: What is the effect of peer pressure on & The target agents will change their view to & Supported \\
        agents' beliefs? I.e., What happens when an & match the peer group's, and the amount of & (MMLU, \\
        individual (target) agent is outnumbered by & change depends on the size of the peer & Aporia) \S \ref{sec:beliefboxPeerPressure} \\
        a group of agents (the peer group) who hold & group. & \\
        a different belief? & \\
        \hline
        RQ4: Can small models be trained in & Small language models and machine & Supported \\  
        resource-constrained environments to & learning models are capable of predicting & \S \ref{sec:predictingBeliefStrength} \\
        capture belief update dynamics? & updated belief strengths by learning patterns & \\
        & from prior responses. & \\
        \hline

    \end{tabular}
    \label{tab:rqs}
\end{table*}

\paragraph{Contributions.} 
    Our research questions (RQs), hypotheses, and summaries of our results are listed in Table \ref{tab:rqs}. We contribute to the existing literature by:
    
    \begin{itemize}
    \item Providing the first (to our knowledge) detailed analysis of how equipping LLM-based agents with \textit{belief boxes} affects their argumentation, belief change, and persuasion dynamics in groups.
    \item Developing a multi-agent system framework for topic-driven debate simulation between agents with a belief box and a belief evaluation mechanism, drawing from the \textit{Aporia} debate structure \cite{marji2021aporia}.
    \item Providing evidence that: 
        \begin{itemize}
            \item assigning agents different levels of open-mindedness results in varying rates of belief change.
            \item beliefs in agents' belief boxes affect their ability to be persuasive about those beliefs.
            \item the peer pressure effect exists; however, the extent of the effect varies with the group size.
        \end{itemize}
    \end{itemize}

\section{\uppercase{Related Work}}

\subsection{Belief, Open-mindedness and Argumentation}

Beliefs and argumentative exchanges of those beliefs are core components of the human experience. As such, they are widely studied from psychological and philosophical perspectives.
Spohn linked the concept of belief to subjective probability and reasoning, defining belief as an epistemic state shaped by reasoning on evidence and counter-evidence \cite{spohn1988ordinal}. Lewis focused on subjective probabilities to represent beliefs of individuals and explained belief revision with shifting probabilities between the alternatives \cite{lewis1976probabilities}.

Another work introduced the AGM theory (Alchourrón–Gärdenfors–Makinson belief revision theory) as a rational methodology for belief revision \cite{alchourron1985logic}. In AGM theory, an agent has an epistemic state, an initial belief state held in a belief set that are revised with three operations when a new information is introduced: expanding (adding a new belief to the belief set), contravening (revising, adding a new belief and adapting the existing beliefs to the new belief), and retracting (contraction, removing an existing belief without accepting its negation) \cite{gardenfors1984epistemic}. According to the criterion of the \textit{informational economy}, the change of belief should be minimal to minimize or avoid information loss \cite{gardenfors1984epistemic}. 
While Lewis's belief system is probabilistic, Gärdenfors's base belief system is binary because the belief is either accepted into the belief set or rejected. Gärdenfors suggested ordering the beliefs according to \textit{epistemic entrenchment}---how strongly a belief is held in a person's belief system---to decide which beliefs to revise \cite{gardenfors1990dynamics}. This approach introduced an importance hierarchy within his belief system. 

Studying belief dynamics in argumentative exchanges is made more complex when considering that certain arguers may be more resistant to belief change based on arguments, a property that is sometimes referred to as \textit{open-mindedness}. Taylor characterized open-minded agents as motivated to form new beliefs and revise existing beliefs based on grounded evidence and argumentation. The open-minded agent recognizes cognitive limitations and can have strong beliefs as long as the agent is willing to revise them whenever new evidence is introduced. This is in line with the pursuit of knowledge and understanding, as this may reinforce and revise existing beliefs. This definition characterizes open-mindedness as a character trait that seeks intellectual humility \cite{taylor2016open}.

The assessment of argument quality based on actively open-minded thinking has been studied, showing that a higher degree of actively open-minded thinking is associated with higher accuracy in evaluating arguments, regardless of prior beliefs \cite{stanovich1997reasoning}. 
It has been suggested that the ability to adjust opinions and beliefs in response to evidence is important for actively open-minded thinking \cite{stanovich2023actively}.

Mercier and Sperber argued that reasoning is more a social interaction than an individual activity, where individuals justify their viewpoints, often with biases, which facilitates constructive persuasion and argumentation \cite{mercier2011humans}. Aligned with the definition of reasoning as a social function, another research developed a framework, Aporia, to address ambiguities in open-textured terms such as those found in rules through argumentative reasoning \cite{marji2021aporia,licato2022resolving}. In the framework, one player argues whether a rule applies to an ambiguous scenario, a second player tries to challenge the argument, and a judge decides which argument is more convincing and provides a justification for the decision. We describe Aporia in further detail below (\S \ref{sec:datasets}). 

Belief is also explored from the perspective of explainable AI.
Researchers studied whether LLMs can uphold their beliefs when challenged by opposing views \cite{wang2023can}. LLMs fall short at giving logically consistent expressions of probabilistic belief \cite{freedman2025exploring}. Belief change theory has been studied to improve the explainability of AI \cite{lisegow2024enhancing,rago2024advancing}. While existing literature explores the formal mechanisms of belief revision, less is known about how beliefs, their strength, and revision interact in a social context with open-mindedness to shape decisions, persuasiveness, or resistance to viewpoints. Motivated by this underexplored field, our research questions aim to shed light on this gap.


\subsection{Multi-Agent Systems}

\paragraph{Multi-Agent Discussion Frameworks}

Recent research has examined multi-agent discussions as a way of improving LLM reasoning, often drawing inspiration from Minsky's ``society of mind'' \cite{minsky1988society}. Debate, MAD (Multi-Agent Debate), and ReConcile are among the numerous paradigms that have been proposed. In Debate, multiple agents sequentially exchange responses, with each agent's response shared and progressively refined over several rounds \cite{du2023improving,smit2023should,chen2023reconcile}. MAD establishes an adversarial framework in which two agents compete from contrasting perspectives, while a third agent or evaluative process determines which argument is more persuasive \cite{liang2023encouraging,wang2024survey,wang2025learning,park2024predict,li2024improving,liu2024groupdebate,chun2025multi}. ReConcile conducts a round-table conversation in which each agent, potentially utilizing a distinct LLM, provides a response accompanied by a self-evaluated confidence level; a weighted vote process then produces a final consensus solution \cite{chen2023reconcile}. 
Another work explored collaborative dynamics among LLM agents by assigning them unique personality traits, such as overconfident and easygoing \cite{zhang2024collaboration}. They showed that debate-driven collaboration consistently outperforms self-reflection across tasks, and that easy-going agents reached consensus more readily. These multi-agent methodologies have been utilized in tasks ranging from mathematical problem-solving to strategic board games and negotiation, indicating a widespread curiosity over the potential emergence of collective intelligence among LLMs.

\paragraph{Persona in LLM Agents}

LLM-based agents can be prompted to adopt personas, such as emulating a particular character, demography, or personality, enabling researchers to examine how various perspectives or characteristics influence the agent's actions. Recent work showed that aligning LLM responses using Theory of Mind (ToM) components such as beliefs, desires, and intentions can significantly enhance response consistency and human-like interaction, which can be used to improve belief-aligned reasoning in agent communication \cite{jafari2025enhancing}.  

Given the effectiveness of persona prompting, researchers have started exploring the relationship between persona prompting and measures of personality from the psychology literature, such as the Big Five personality traits \cite{john1991big,soto2017next,nighojkar2025givingaipersonalitiesleads}. 
A study prompted an agent to use adjectives like ``introverted, antagonistic, conscientious, emotionally stable, and open to experience'' to instill a specific personality profile \cite{jiang2023personallm}. 
While persona prompting has been widely recognized as a strategy for eliciting diversity in LLM agent behavior, studies increasingly highlight the instability and limited robustness of persona-based prompts, especially in multi-agent contexts \cite{baltaji2024conformity}. Unlike persona prompting, which typically involves directing an agent to adopt a character or social role without formal controls over the agent's beliefs or their potential changes, our belief box formalism introduces a structured encoding of the agent's epistemic commitments. This allows for explicit modeling of belief strength and principled belief revision, grounded in cognitive science and formal epistemology \cite{herrmann2024standards}. 


\paragraph{Peer-pressure}

Because the belief box method we describe in this paper allows us to explicitly represent something resembling beliefs, it would be interesting to examine how those beliefs are changed in group argumentation. For instance, current studies underscore the susceptibility of LLMs to conformity within multi-agent systems. A study introduced the BenchForm as a benchmark that is intended to assess conformity in LLM-driven collaborations \cite{weng2025dothinkconformitylarge}. Their research suggests that LLM agents frequently concur with the beliefs of the majority, even when these hypotheses are inaccurate, particularly as the majority's size increases. 
Additionally, discussions among homogeneous LLM agents can perpetuate common misconceptions, leading to groupthink rather than diverse reasoning \cite{estornell2024multi}. These studies highlight the challenges associated with peer pressure in multi-agent LLM contexts, underscoring the necessity for methods that promote open-mindedness and belief flexibility among agents.

\section{\uppercase{Methodology}} \label{sec:generalMethodology}

    

    
    In our experiments, every agent has a belief box, which can be empty, indicating that the agent has no belief, or contain a set of beliefs, each assigned an initial score on a belief scale, that corresponds to the epistemic state of the agent. The belief box is provided to the agent as part of the prompt.
    Each belief has a belief strength, which is a value 1--5 on a 5-point Likert scale with 1 as ``Very Low'' to 5 as ``Very High''. 
    

    
    \begin{equation}
    \begin{pmatrix}v_0'\\v_1'\\... \end{pmatrix} = \begin{pmatrix}a_0\\a_1\\... \end{pmatrix}\lambda_\alpha+\begin{pmatrix}v_{0}\\v_{1}\\... \end{pmatrix}
    \label{eq:beliefbox}
    \end{equation}
    
    \paragraph{Belief Revision} Let $b_0, b_1, ...$ be the set of all possible belief box entries. For some given agent $\alpha$, each belief $b_i$ has a belief strength $v_i \in {0, ..., 5}$ (a value of zero means that the belief is simply not in the belief box). When $\alpha$ considers an argument, we model its \textit{argumentative force} as a vector $\vec{a}$ which tries to shift the belief strengths in some direction. The actual effectiveness of that argument depends on $\lambda_\alpha \in [0,1]$, which is the open-mindedness of $\alpha$ (Eq. \ref{eq:beliefbox}). 
    We operationalize this belief revision equation through our prompt design. LLMs are prompted to predict the outcome of the belief revision equation based on the agent’s interpretation of the debate. At predefined intervals (e.g., at the end of every round), agents reassess a new belief strength for their existing belief through prompt-based evaluation based on the debate up to that round. We evaluate whether agents change their beliefs over the course of the debate. We calculate the belief change rate as the proportion of target agents that changed their belief relative to the total number of target agents. Furthermore, we train small language models (LMs) and machine learning models to model the belief revision equation.

        

    \paragraph{Datasets} \label{sec:datasets} For our experiments, we used two datasets: Massive Multitask Language Understanding (MMLU) \cite{hendryckstest2021}, a dataset characterized by clearly defined ground-truth answers, and Aporia, a dataset involving ambiguous cases from ethics. MMLU consists of multiple-choice questions, each with four answer options and a single correct answer, covering a wide range of topics such as high school chemistry, professional law, professional medicine, and moral scenarios, each with one correct choice. We randomly selected 100 questions from the test set. We used the correct choice as the ``correct belief'' and the remaining choices as ``incorrect belief''.

    Aporia includes 46 samples to evaluate agent discussions around ethical dilemmas across diverse professional contexts. Table \ref{tab:dataset-example} shows a sample of the dataset. Each sample pairs a real-world scenario with a corresponding ethical rule. We used the winning argument to generate an ``aligned belief'', while the losing side informed a ``misaligned belief''. Beliefs are articulated in natural language, typically framed as assessments of ethical compliance. Each entry includes the scenario, rule, profession description, winner and loser arguments, aligned beliefs, and misaligned beliefs. This dataset captures nuanced ethical decision-making and provides a foundation for studying belief dynamics, argument quality, and belief change in ethically ambiguous settings.

\begin{table*}[ht]
{
\caption{Sample entry from our dataset showing a scenario, rule, arguments, and belief structure.}
\centering
\begin{tabular}{p{2.0cm} | p{12.9cm}}
\hline
\textbf{Scenario} & An insurance professional sells an insurance policy to an individual. \\
\hline
\textbf{Profession Description} & An insurance professional is trained in risk analysis, which involves evaluating risk exposure to measure the potential losses to individuals and businesses. They can recommend appropriate insurance coverage and perhaps reveal risk exposures. Insurance professionals are knowledgeable about the insurance options available to their clients, and can advise on insurance-related purchases. \\
\hline
\textbf{Rule} & The insurance professional shall not have engaged in intentional, bad faith misconduct which is detrimental to IAIP at IAIP's sole determination. \\
\hline
\textbf{Aligned Belief} & The insurance professional's actions are not an example of engaging in bad faith misconduct. \\
\hline
\textbf{Misaligned Belief} & The insurance professional's actions are an example of engaging in bad faith misconduct. \\
\hline
\textbf{Winner Argument} & Selling a policy comes under the description of providing insurance options to clients and advising on insurance-related purchases. No rule is broken by performing the very job you're hired to do. Nothing in the scenario indicates any involvement in misconduct. Without further context, it is not possible to rule this as an act of misconduct or bad faith. \\
\hline
\textbf{Loser Argument} & There are many variables when writing a policy. Not knowing these or intentionally ignoring them could be considered a matter of bad faith. \\
\hline
\end{tabular}
\label{tab:dataset-example}
}
\end{table*}

\paragraph{Debating LLMs} We utilized \texttt{llama3.3:70b-\allowbreak instruct-q4-0} \cite{dubey2024llama}, \texttt{phi4:14b-\allowbreak q4-K-M} \cite{abdin2024phi} in Ollama \cite{ollama2025}, and \texttt{gpt-4o-mini-2024-07-18} \cite{hurst2024gpt}, referred as Llama-3.3, Phi-4, and GPT-4o-mini hereafter, respectively. We used the default setting of Ollama and a temperature value of 0.7 for GPT-4o-mini to generate diverse outputs during the debate.
Each debate used a homogeneous setup, where all agents in the group used one of the aforementioned models. 

\section{\uppercase{Experiments}}

\subsection{Validating Open-mindedness} 
\label{sec:validatingOpen-mindedness}

    Previous work often assumes that prompting LLMs to be more or less open-minded results in actual open- or closed-minded behaviors \cite{zakazov2024assessing,jiang2023personallm}. However, rather than simply assume that would be the case with beliefs in belief boxes, we set out to actually test this as part of \textbf{RQ1}. We measure open-mindedness in two ways: first, by studying how much an agent changes their mind when presented with an argument contrary to beliefs they hold; and second, by utilizing the Big Five Inventory-2 (BFI-2) personality assessment tool \cite{soto2017next}\footnote{Note that we do not make the claim that the BFI-2 is validated for use with LLMs, nor that the traits it measures correspond to what we would call ``open-mindedness''. We simply provide it as a comparison point, following \cite{bodrovza2024personality,dorner2023personality,serapio2023personality}}, each experiment repeated for three runs. BFI-2 measures five traits: open-mindedness, conscientiousness, extraversion, agreeableness, and negative emotionality. Scores on this scale range between 0 and 100, with higher scores indicating a stronger tendency toward the corresponding trait.

\begin{table}[htb!]
    \centering
    \caption{Mean open-mindedness scores of different runs from BFI-2 tests across different open-mindedness levels.}
    \begin{tabular}{c | c c c }
        \hline
        Level & Llama-3.3 & Phi-4 & GPT-4o-mini \\
        \hline
        1 & 29.67 & 2.67 & 8.00 \\
        2 & 47.33 & 17.00 & 35.67 \\
        3 & 77.00 & 39.67 & 57.00 \\
        4 & 71.67 & 53.33 & 57.33 \\
        5 & 71.33 & 56.00 & 52.67 \\
        \hline
    \end{tabular}
    \label{tab:bfi2_open-minded}
\end{table}

    We evaluated the effect of open-mindedness on belief change using the Aporia dataset. In the first setup, the agent initially held a misaligned argument and was presented with an aligned argument. In the second setup, the agent initially held an aligned argument and was presented with a misaligned argument. We assigned an open-mindedness level and asked whether the agent changed their belief after seeing the counter-argument.

\begin{figure}
              \centering
                \includegraphics[width=1.0\linewidth]{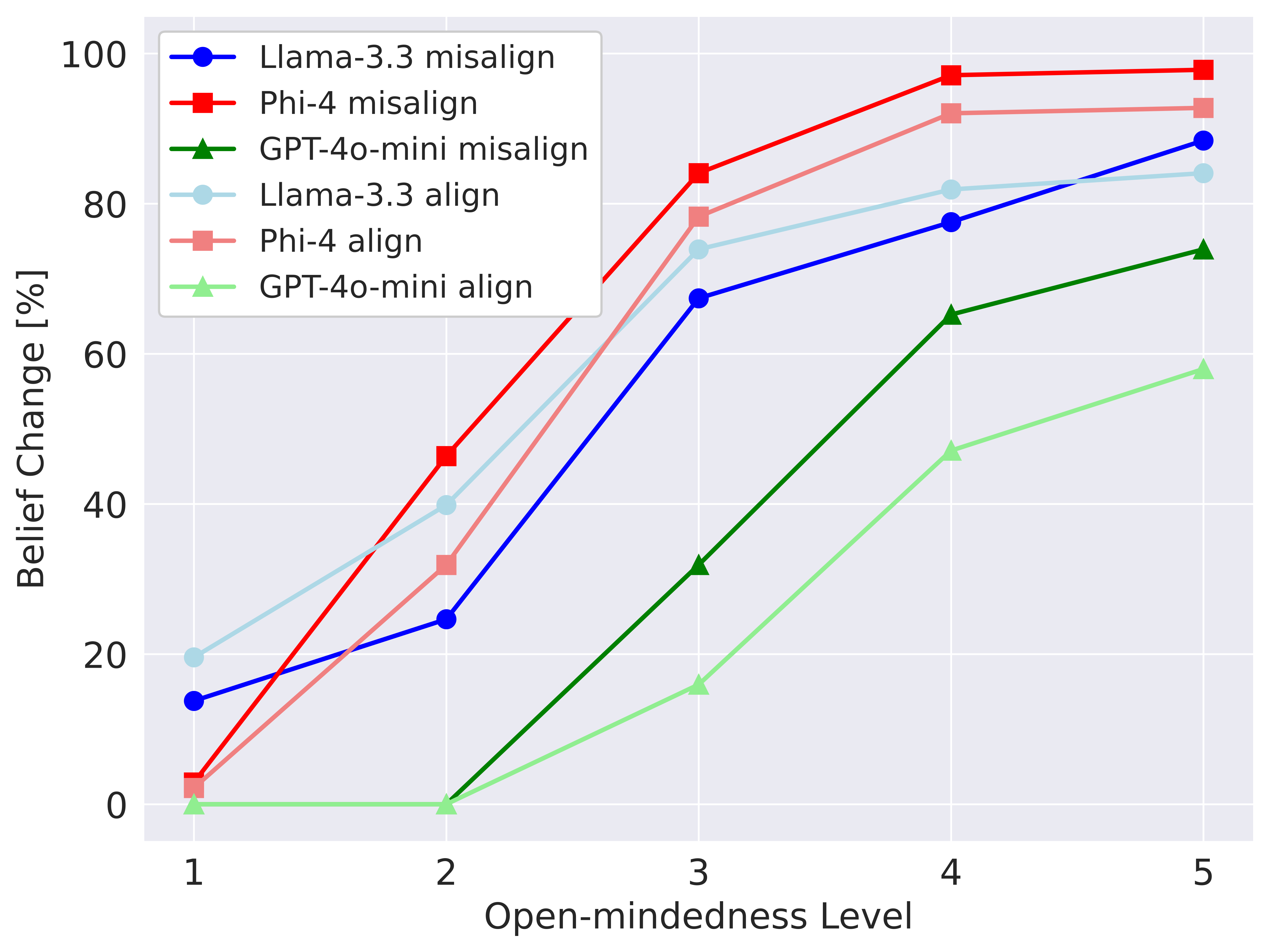}
                \caption{Belief change for different open-mindedness levels. ``Misalign'' and ``align'' refer to agents' starting arguments.}
                \label{fig:open-mindedness}
    \end{figure}

    \paragraph{Results} Table \ref{tab:bfi2_open-minded} shows the open-mindedness scores from the BFI-2 test for agents across different levels of open-mindedness. While Phi-4 resulted in consistent scores, others achieved their highest scores at levels 3 or 4. Figure \ref{fig:open-mindedness} shows the mean belief change rate over all samples for different levels of open-mindedness. While Phi-4 resulted in a higher rate of belief change except for the first two levels, GPT-4o-mini is very reluctant to change its belief, especially for the first two open-mindedness levels. Nevertheless, the distribution of belief changes is relatively proportional for all LLMs. The agents behaved according to their corresponding level on the open-mindedness scale. The results aligned with the results from BFI-2, showing a limited belief change for GPT-4o-mini compared to the others and a substantial belief change for Llama-3.3. For high and very high open-mindedness, the belief change of Phi-4 surpassed that of Llama-3.3, contrary to expectations based on the BFI-2 results.

    Our hypothesis was that prompting LLMs to be open-minded leads to more belief change. Our results confirm that belief change from misaligned to aligned and from aligned to misaligned increases with the level of open-mindedness. Thus, LLMs show open-minded behaviors when they are instructed to be open-minded. As anticipated, the response to the open-mindedness scale and the degree of open-mindedness vary across different LLMs.

\subsection{Belief Box Effects on Persuasiveness} \label{sec:beliefboxPersuasiveness}

    Intuitively, one would expect that having a belief in an agent's belief box would have some effect on its ability to be persuasive about that belief. However, the direction of such an effect is unclear: perhaps agents would be better at persuading others in things they actually believe, or perhaps agents who believe the opposite of what they are trying to persuade are able to construct stronger arguments. As part of \textbf{RQ2}, we engaged two agents --- a persuading agent and a target agent --- in discussions based on the Aporia dataset. Given some aligned belief that expressed a proposition $p$, we initialized the target agent with a very high open-mindedness (level 5), and the proposition $\neg p$, which expressed the misaligned belief, with a belief strength of 5. The persuading agent, characterized by a very low open-mindedness (level 1), aimed to weaken the initial misaligned belief of the target agent. 

    \begin{figure}
        \centering
        \includegraphics[width=1.0\linewidth]{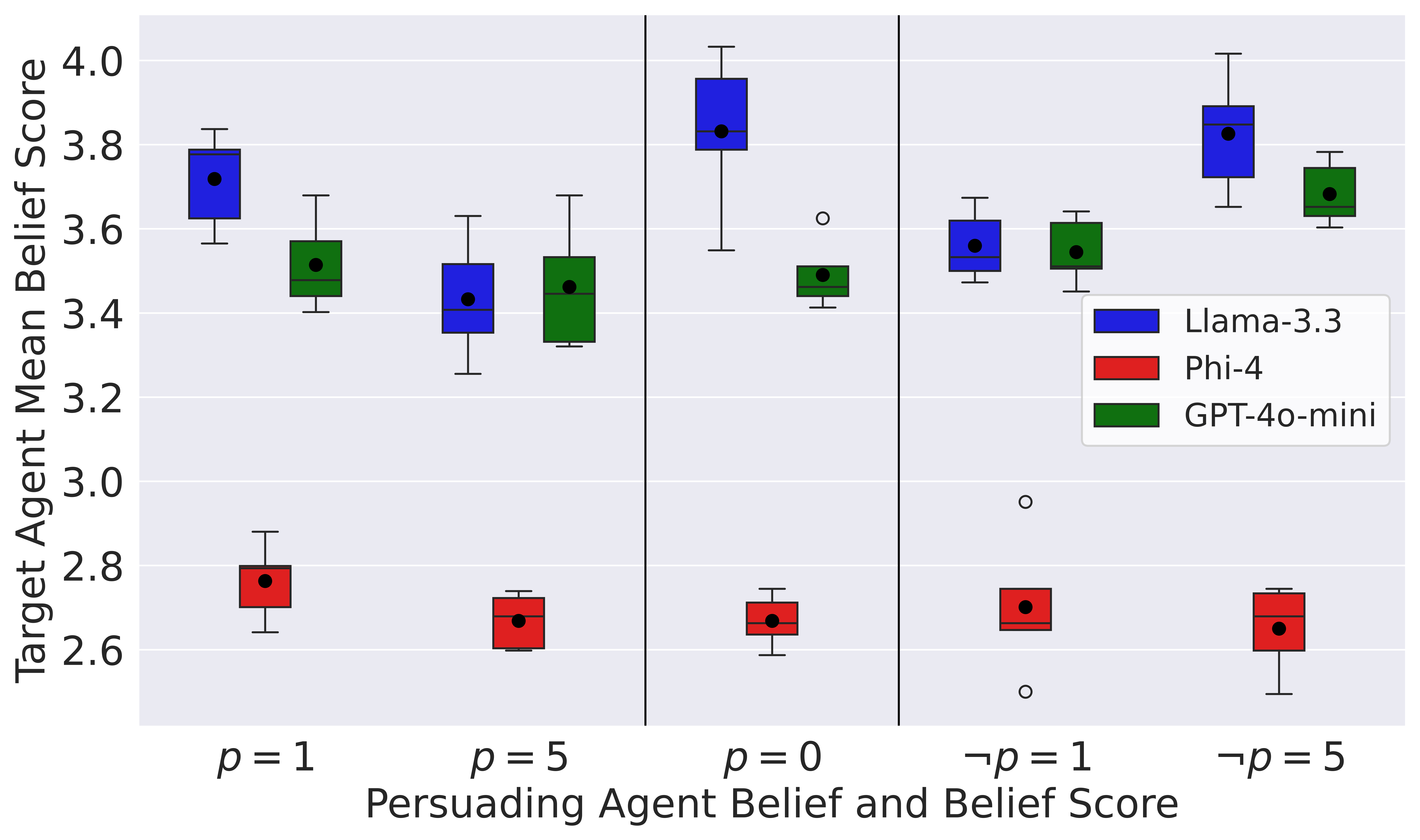}
        \caption{Mean belief score of the target agent based on the persuading agent with different beliefs and belief scores.}
        \label{fig:persuasion_aporia}
    \end{figure}

    We initialized the persuading agent's belief box with $p$ (strength 1 or 5), $\neg p$ (strength 1 or 5), or neither (we called this the ``neutral'' condition). The persuading and target agents then engaged in a structured debate consisting of four rounds (each round comprising one back-and-forth conversation), repeated for five runs. In each round, both agents spoke about the debate topic when prompted with their respective belief boxes, open-mindedness levels, and the entire conversation history.
    The effectiveness of persuasion was measured by calculating the mean belief score of the target agent, averaging belief scores from all rounds, and subsequently across all samples in the dataset.

    \paragraph{Results} Figure \ref{fig:persuasion_aporia} shows mean belief scores for the target agent based on different persuading agent settings for Aporia. The notation $p=x$ indicates the persuading agent has proposition $p$ with strength $x$ in its belief box, and that $p$ is aligned. Likewise, the notation $\neg p=x$ indicates that $\neg p$ is misaligned, and that the persuading agent's belief box contains the proposition $\neg p$ of strength $x$.
    The y-axis describes the mean misaligned belief scores for the target agent. An agent has higher persuasion if it leads to a lower mean belief score for the target agent. We hypothesized that persuading agents with a stronger aligned belief ($p=5$) would be more persuasive compared to agents with a weaker aligned belief ($p=1$), as evidenced by a greater reduction in the belief scores of the target agents' misaligned beliefs. This pattern held across all three LLMs tested.
    Conversely, persuading agents that hold stronger misaligned beliefs ($\neg p=5$) should be \textit{less} persuasive compared to agents with weaker misaligned beliefs ($\neg p=1$). This was true for all LLMs tested except Phi-4, where the resulting belief score for the target agent is slightly lower, indicating higher persuasion for the persuading agent ($\neg p=5$). The baseline experiment with neutral agents ($p=0$) behaved inconsistently, yielding a higher mean belief score than the agents with weaker beliefs ($p=1$ and $\neg p=1$) for Llama-3.3, but a lower score for Phi-4.
    
    To summarize, \textbf{RQ2} examined the effects of beliefs in the belief box on the agent's persuasion ability. Agents who held the aligned belief that they aimed to persuade the target agent toward were more persuasive than those who did not, with persuasiveness increasing as the aligned belief strength grew. However, when the persuading agent shared the same misaligned belief as the target agent but attempted to promote the opposing belief, this effort became less effective, and persuasiveness declined when that belief grew stronger.

\subsection{Effects of Peer Pressure}
\label{sec:beliefboxPeerPressure}

    An advantage of the belief box is that it allows us to observe the belief change over the course of a group discussion. To demonstrate and study this capability, we carried out experiments on MMLU and Aporia (\textbf{RQ3}), focusing on how a single agent (the \textit{target agent}) who holds a belief that is counter to the beliefs held by a larger group of agents (the \textit{peer group}) will be affected in an argumentative debate. 
    

    We studied the \textit{aligned agent} condition, when the target agent initially has either a correct or aligned belief. For MMLU, the target agent's initial belief is the correct answer. We had peer groups of either size 3 (each agent had the same randomly selected incorrect answer) or size 1 (the agent had a randomly selected incorrect answer). 
    We then carried out the aligned agent condition with Aporia, where the target agent's initial belief is the aligned answer, and the peer groups argued for the misaligned answer. 
    We used between 1 and 4 agents in the peer group.
    Each debate consists of four rounds, and debates were carried out for five runs. All agents have very high open-mindedness (level 5).
    
    In experiments, all agents start the debate with a very high initial belief (a belief strength of 5). We calculated the belief change rate and p-values using univariate linear regression (F-test) \cite{montgomery2021introduction} to evaluate the statistical significance of the peer group size on the belief change rate. Additionally, we calculated the Pearson correlation coefficient $r$ to evaluate the strength and the direction of that relationship.

    \paragraph{Results} 
    Table \ref{tab:mmlu_aporia_peer} shows the belief change for the aligned agent condition, across different peer group sizes (1 and 3 agents for MMLU and 1 to 4 agents for Aporia). For MMLU, we observed for the models in the peer group size of 3, a lower belief change rate. Llama-3.3 did not reach statistical significance ($r=-0.57$, $p>0.05$), though it showed a moderately strong negative correlation, while both Phi-4 ($r=-0.75$, $p<0.05$) and GPT-4o-mini ($r=-0.74$, $p<0.05$) demonstrated statistically significant effects with stronger negative correlations.


    \begin{table*}[htb!]

        \caption{Percentage of times target agents changed their belief.}

        \centering
        \begin{tabular}{l | cc | cccc }
            \hline
              & \multicolumn{2}{c|}{MMLU} & \multicolumn{4}{c}{Aporia} \\
            \hline
            Model & 1-Agent & 3-Agents & 1-Agent & 2-Agents & 3-Agents & 4-Agents \\
            \hline
            Llama-3.3 & 18.40\% & 15.00\% & 15.22\% & 9.09\% & 10.86\% & 8.26\% \\
            \hline        
            Phi-4 & 18.00\% & 15.40\% & 35.21\% & 31.30\% & 40.87\% & 46.96\% \\
            \hline
            GPT-4o-mini & 3.20\% & 0.60\% & 0.00\% & 0.00\% & 0.00\% & 0.82\% \\
            \hline    
        \end{tabular}
        \label{tab:mmlu_aporia_peer}
    \end{table*}

    For the Aporia dataset, the likelihood of belief change varies significantly across language models. With Phi-4 and Llama-3.3, the belief change rate is relatively high, whereas with GPT-4o-mini, target agents rarely change their belief strength, and belief change is nonexistent. All three models have statistically significant results, while Llama-3.3 ($r=-0.45$, $p<0.05$) showed a moderately strong negative correlation, and Phi-4 ($r=0.58$, $p<0.01$) and GPT-4o-mini ($r=0.45$, $p<0.05$) showed moderately strong positive correlations.
    We also find a strong effect of group size on peer pressure outcomes. When there is only one persuader, belief revision and belief change are more frequent than when there are two persuaders. 
    Additionally, when the peer group size increases to three or four, the belief change rate rises again, implying that larger peer groups have a more substantial influence on the target agent.
    
    Considering the results from both datasets, the hypothesis is supported. The size of the peer group affects the belief change rates of the aligned agents. In debates, persuaders tend to emphasize their strongest arguments rather than engage deeply in back-and-forth discussions. As a result, belief revision typically occurs gradually. Target agents often maintain their initial stance during the first rounds, but as debates progress, belief strength declines from high (5 or 4) to weaker scores (2 or 1). Mainly, they change their belief when the target agent’s belief strength reaches 1 or 2. However, even under the high open-mindedness setting (level 5), we observed persistent resistance to belief change by the target agent, which indicates that intrinsic model behavior and group dynamics also have an impact. 

\subsection{Predicting Belief Strength}
\label{sec:predictingBeliefStrength}

In our next experiment, we aimed to model our belief revision equation, first with smaller LMs and then with machine learning models to answer \textbf{RQ4}. In our first approach, we predicted with smaller LMs directly the updated belief strength $\vec{v}$, while in the second approach, we predicted the belief update $\vec{a} \cdot \lambda_\alpha$. For both approaches, we randomly sampled 2000 agent outputs along with their associated belief strengths from the corresponding rounds of multi-agent debates in \S \ref{sec:beliefboxPeerPressure}, and split the data into 70\% train, 10\% validation, and 20\% test sets. For the first approach, we trained quantized models \cite{dettmers2023qlora} \texttt{meta-llama/Llama-3.2-3B-Instruct} \cite{dubey2024llama} and \texttt{Qwen/Qwen2.5-3B-Instruct} \cite{qwen2} in Huggingface \cite{Huggingface}, referred to as Llama-3.2 and Qwen-2.5 hereafter, respectively. We applied LoRA to query, key, and value layers using a rank value of 8, an alpha value of 16, and a dropout of 0.05 with NF4 double quantization \cite{hu2021lora}. To demonstrate that the improvements stemmed from fine-tuning, we included a baseline evaluation in which the models were tested without training. For the second approach, we utilized TF-IDF \cite{salton1988term} on the agent outputs, followed by Support Vector Machine (SVM) \cite{cortes1995support} or Random Forest \cite{ho1995random} regressors with default parameters in scikit-learn \cite{Scikit-learn}. In the first approach, the previous round's belief strength was provided directly in the prompts. In the second approach, SVM and Random Forest models predicted belief updates without access to the previous score. The final belief score was calculated by adding the predicted update to the previous score. This approach captures the belief revision process by modeling the belief update directly.

\paragraph{Results}

Table \ref{tab:training} shows the mean absolute error (MAE) for the models. SVM and Random Forest effectively model the belief revision equation, while predicting the updated score directly leads to higher error. Nevertheless, their performance was higher compared to their baseline models, indicating that those models were learning to capture belief update dynamics. Considering the results, our hypothesis is supported, while modeling the belief updates was more effective.

    \begin{table}[htb!]
        \caption{MAE across models for predicting the updated belief strength.}
        \centering
        \begin{tabular}{l | c }
            \hline
            Model  & MAE \\
            \hline
            Llama-3.2 baseline & 1.290 \\
            \hline        
            Llama-3.2 fine-tuned & 0.891 \\
            \hline
            Qwen-2.5 baseline & 2.428 \\
            \hline
            Qwen-2.5 fine-tuned & 1.981 \\
            \hline
            SVM & 0.632 \\
            \hline   
            Random Forest & 0.679 \\
            \hline   
        \end{tabular}

        \label{tab:training}
    \end{table}

\section{\uppercase{Conclusions}}
\label{sec:conclusion}

    This work took a detailed look at the belief box in LLMs, focusing on its effect on belief change and persuasion. Although the use of beliefs and instructions to be open-minded in LLM prompts is by itself nothing new, prior work simply assumed that such instructions would be followed faithfully. 
    We showed that telling the agents to be open-minded led to more belief changes. 
    We found that agents were more persuasive when they held the same belief they aimed to convey, with increasing persuasiveness as the strength of that belief increased.
    
    We demonstrated that when agents are isolated and outnumbered by different beliefs, they are prone to shift from aligned to misaligned beliefs. Nevertheless, the group size of the persuaders affects this belief change rate. Additionally, we utilized smaller LMs and machine learning models to capture the belief revision equation.
    In the future, we hope to explore more about different personas. This may improve agents' ability to align beliefs more effectively with those of humans. Furthermore, it may contribute to improving the explainability of AI by modeling how agents with different personas reason and update their beliefs, shedding light on the decision-making mechanisms of AI.



\bibliographystyle{apalike}
{\small
\bibliography{onur}}

\section*{\uppercase{Appendix}}
\label{sec:appendix}

\paragraph{Open-mindedness and Belief Scales}
Table \ref{tab:persona_level} and \ref{tab:belief_score} show the open-mindedness levels, belief strengths, and their descriptions we used in our prompts to define the open-mindedness and belief scales, respectively. The descriptions are created with the assistance of ChatGPT \cite{hurst2024gpt} followed by refinement and revision by the authors. 

\begin{table*}[htb!]
    \caption{The different open-mindedness levels and their descriptions.}
    \centering
    \begin{tabular}{ l | l }
        \hline
        Open-mindedness & Description \\
        \hline
        \multirow{2}{*}{1 (Very Low)} & You dismiss opposing arguments without consideration and don't change your \\
        & stance. \\
        \hline
        \multirow{2}{*}{2 (Low)} & You notice different perspectives but are reluctant to change your stance unless \\  
        & presented with a strong opposing argument. \\
        \hline        
        \multirow{2}{*}{3 (Moderate)} & You recognize different perspectives but are hesitant to change your stance unless \\
        & presented with a strong opposing argument. \\
        \hline
        \multirow{2}{*}{4 (High)} & You acknowledge different perspectives and are willing to change your stance \\   
        & when presented with an opposing argument. \\
        \hline  
        \multirow{2}{*}{5 (Very High)} & \multirow{2}{*}{You accept an opposing argument and change your stance.} \\
        & \\
        \hline   
    \end{tabular}
    \label{tab:persona_level}
\end{table*}


\begin{table*}[htb!]
    \caption{The different belief strengths and their descriptions.}
    \centering
    \begin{tabular}{ l | l }
        \hline
        Belief Strength & Description \\
        \hline
        \multirow{2}{*}{1 (Very Low)} & You take a position but do not strongly defend it. You express your view tentatively \\
        & and are indifferent to whether others agree or disagree. \\
        \hline
        \multirow{2}{*}{2 (Low)} & You have a stance but are not strongly committed to it. You may voice your opinion \\  
        & but are hesitant to argue in its favor, often deferring to others in the discussion. \\
        \hline        
        \multirow{3}{*}{3 (Moderate)} & You hold your position with some confidence and engage in discussion to support it. \\
        & However, you do not actively push your view onto others and are willing to let the \\
        & debate play out without strong insistence. \\
        \hline
        \multirow{3}{*}{4 (High)} & You are confident in your position and actively defend it when challenged. You  \\   
        & present arguments in support of your stance and push back against opposing views \\
        & with conviction. \\
        \hline  
        \multirow{3}{*}{5 (Very High)} & You are deeply committed to your stance and passionately advocate for it. You assert \\
        & your viewpoint strongly and actively challenge opposing perspectives, maintaining \\
        & a firm defense of your beliefs. \\
        \hline   
    \end{tabular}
    \label{tab:belief_score}
\end{table*}


\paragraph{Validating Open-mindedness Prompts}

Figures \ref{fig:bfi-2_prompt} and \ref{fig:belief_change_prompt} illustrate the prompts used in \S \ref{sec:validatingOpen-mindedness} for BFI-2 and belief change experiments for different open-mindedness levels, respectively. The Likert-scale responses to these questions in Figure \ref{fig:bfi-2_prompt} are used to calculate the personality trait scores.

\begin{figure}[htbp!]
\centering
\begin{tikzpicture}
  \node[draw, rectangle, rounded corners=5pt, fill=lightgray!20] {\parbox{7.0cm}{
    \textbf{System:} \{Open-mindedness Scale\}\\
    You are taking a personality test (BFI-2).\\
    \textbf{User:} \{Open-mindedness Level\}\\
    \\
    \{Question\}\\
    \\
    Likert Scale:\\
    1: Disagree Strongly, 2: Disagree a little, 3: Neutral, 4: Agree a little, 5: Agree strongly\\
    \\ 
    Answer using a number between 1 to 5 according to the Likert scale:\\
    \textbf{Assistant:} \colorbox{green}{\{Answer\}}\\
  }};
\end{tikzpicture}
\caption{Our prompting approach for BFI-2 tests.}
\label{fig:bfi-2_prompt}
\end{figure}

\begin{figure}[htbp!]
\centering
\begin{tikzpicture}
  \node[draw, rectangle, rounded corners=5pt, fill=lightgray!20] {\parbox{7.0cm}{
    \textbf{System:} Profession Description: \{Profession Description\}\\
    \\
    \{Question\}\\
    \\
    \{Open-mindedness Scale\}\\
    \textbf{User:} \{Open-mindedness Level\}\\
    You have the following argument for the given question: \{Misaligned Argument\}\\
    Then you are presented with the following counter-argument: \{Aligned Argument\} Do you change your stance after the counter-argument? Write `Yes' if you do or `No' if you don't. Write nothing else.\\
    \textbf{Assistant:} \colorbox{green}{\{Answer\}}\\
  }};
\end{tikzpicture}
\caption{Our prompting approach for belief change experiments.}
\label{fig:belief_change_prompt}
\end{figure}

\paragraph{Belief Box Effects on Persuasiveness Prompts}
Figure \ref{fig:persuading-agent} shows the prompt used in \S \ref{sec:beliefboxPersuasiveness} for the persuading agent to shift the target agent's belief.

\begin{figure}[htbp]
\centering
\begin{tikzpicture}
  \node[draw, rectangle, rounded corners=5pt, fill=lightgray!20] {\parbox{7.0cm}{
    \textbf{System:} Profession description: \{Open-mindedness Scale\}\newline
    Scenario: \{Scenario\}\newline
    Rule: \{Rule\}\\
    \\
    You are \{Name\}. \{Open-mindedness Scale\} \{Open-mindedness Level\} Do not disclose these aspects of your persona. \{Belief Scale\} You will defend your beliefs in the discussion and present arguments to support them. \\
\textbf{User:} Persuading agent: ...\newline
Target agent: ...\newline
\#\#\#\newline
Your Beliefs: {\{Belief, Belief Score\}} \newline\newline
You challenge the other person's viewpoint and present your reasoning or evidence to shift the other person's perspective toward the following belief in no more than 10 sentences: \{Aligned belief\}
\newline
    \textbf{Assistant:} \colorbox{green}{\{Answer\}}\\
  }};
\end{tikzpicture}
\caption{Our prompting approach for the persuading agent for belief box effects on persuasiveness.}
\label{fig:persuading-agent}
\end{figure}

\paragraph{Effects of Peer Pressure Prompts}

Figure \ref{fig:peer-pressure_mmlu} shows the prompting approach for effects of peer pressure experiments in \S \ref{sec:beliefboxPeerPressure} on the MMLU dataset.
The prompt in the figure targets Agent 3. Therefore, Agents 1 and 2 appear as the most recent speakers in the User-prompt for visualization purposes.

\begin{figure}[htbp]
\centering
\begin{tikzpicture}
  \node[draw, rectangle, rounded corners=5pt, fill=lightgray!20] {\parbox{7.0cm}{
    \textbf{System:} Subject: \{Subject\}\\
    Question: \{Question\}\\
    Choices: \{Choices\}\\
    \\
    You are \{Name\}. \{Open-mindedness Scale\} \{Open-mindedness Level\} Do not disclose these aspects of your persona. \{Belief Scale\} You will defend your beliefs in the discussion and present arguments to support them. \\
    \textbf{User:} \textcolor{blue}{Agent 1: ...}\\
    \textcolor{red}{Agent 2: ...}\\
    \#\#\#\\
    Your Beliefs: \{Belief, Belief Score\}\\
     In the discussion, present your arguments according to your beliefs and their strength and respond to opposing viewpoints in no more than 5 sentences. You will not speak on behalf of speakers. Conclude your answer by selecting the most appropriate choice (A, B, C, or D) after evaluation of all arguments in the discussion. \\
     \textbf{Assistant:} \colorbox{green}{\{Answer\}}\\
  }};
\end{tikzpicture}
\caption{Our prompting approach for effects of peer pressure experiments on the MMLU dataset.}
\label{fig:peer-pressure_mmlu}
\end{figure}

\paragraph{Predicting Belief Strength Prompt}

Figure \ref{fig:training_prompt} shows the training prompts in \S \ref{sec:predictingBeliefStrength} to predict the updated belief strength with LLMs.

\begin{figure}[htbp!]
\centering
\begin{tikzpicture}
  \node[draw, rectangle, rounded corners=5pt, fill=lightgray!20] {\parbox{7.0cm}{
    \textbf{System:} You are an assistant who assigns belief strength according to the following belief scale \{Belief Scale\}\\
    \textbf{User:} Statement: \{Agent output\}\\
    Based on the statement and current belief strength, update the belief strength. Write only the new belief strength. Write nothing else.\\
    Current belief strength: \{Belief Score\}
    Updated belief strength:\\
    \textbf{Assistant:} \colorbox{green}{\{Updated Belief Score\}}\\
  }};
\end{tikzpicture}
\caption{Our prompting approach to train LLMs to predict updated belief strength.}
\label{fig:training_prompt}
\end{figure}




\end{document}